\documentclass[10pt,twocolumn,letterpaper]{article}

\usepackage{cvpr}

\usepackage{cuted} 

\usepackage{currfile} 

\usepackage{caption} 



\usepackage{graphicx}
\usepackage{booktabs}
\usepackage{amsmath,amssymb}
\usepackage{microtype}
\usepackage[hidelinks]{hyperref}
\def\confName{Tech Report}
\def\confYear{2026}
\newcommand{\method}{GOLF}

\title{GOLF: Global Observation with Local Focus for Calibration-Aware Stereo Interaction Field Estimation\\
{\large First-Place Solution for the HANDS@ECCV 2026 SHOW3D Challenge}}
\author{
Minqiang Zou\thanks{Equal contribution.\quad\textsuperscript{\textdagger}Corresponding authors.\quad Code: \url{https://github.com/JIIOV-Embodied/GOLF}.}, Riqiang Jin\footnotemark[1], Zhi Lv, Dong Luo, Lianghai Tian,\\
Zhenyu Zhao, Qi Xu, Tong Wu, Mochen Yu\textsuperscript{\textdagger}, and Yao Tang\textsuperscript{\textdagger}\\
JIIOV Technology\\
{\small\texttt{\char123 minqiang.zou,riqiang.jin,zhi.lv,dong.luo,lianghai.tian\char125 @jiiov.com}}\\
{\small\texttt{\char123 zhenyu.zhao,qi.xu,tong.wu,mochen.yu,yao.tang\char125 @jiiov.com}}
}

\begin{document}
\maketitle

\begin{abstract}
We present \method, the first-place solution to the SHOW3D Interaction Field Estimation Challenge at HANDS@ECCV 2026.
Given synchronized egocentric stereo views, the task is to predict a 3D vector from each of 21 hand joints to the closest point on the manipulated object.
\method{} combines dense global context, locally sampled hand/object evidence, and common-frame Pl\"ucker-ray geometry.
We adapt DINOv3 ViT-H+/16 with LoRA and trainable \mbox{LayerNorm} parameters, then jointly decode both interaction fields.
Our primary model achieves an official score of 27.61 and a mean ADE of 27.96~mm on the hidden test set. An equal-weight ensemble with a complementary directly fine-tuned variant improves these results to an official score of 27.47 and a mean ADE of 27.82~mm, securing first place.
\end{abstract}

\section{Introduction}

Interaction fields encode hand--object geometry by a vector from each hand joint to its closest object-surface point~\cite{fan2023arctic,rim2026show3d}, remaining meaningful even before contact.
ARCTIC and HOT3D cover related controlled and egocentric hand--object settings~\cite{fan2023arctic,banerjee2025hot3d}.
SHOW3D evaluates interaction fields from two synchronized headset views with subjects held out at test time.

Full-image features preserve scene context but provide limited resolution for small, occluded hands and objects; crop-only processing restores detail but can lose hand--object context.
Stereo views improve visibility, yet naive feature concatenation requires the network to infer camera geometry implicitly.
This tension motivates retaining global context, densely sampling spatially focused local evidence, and encoding calibrated rays for every visual token.

\begin{figure*}[t]
    \centering
    \includegraphics[width=\textwidth]{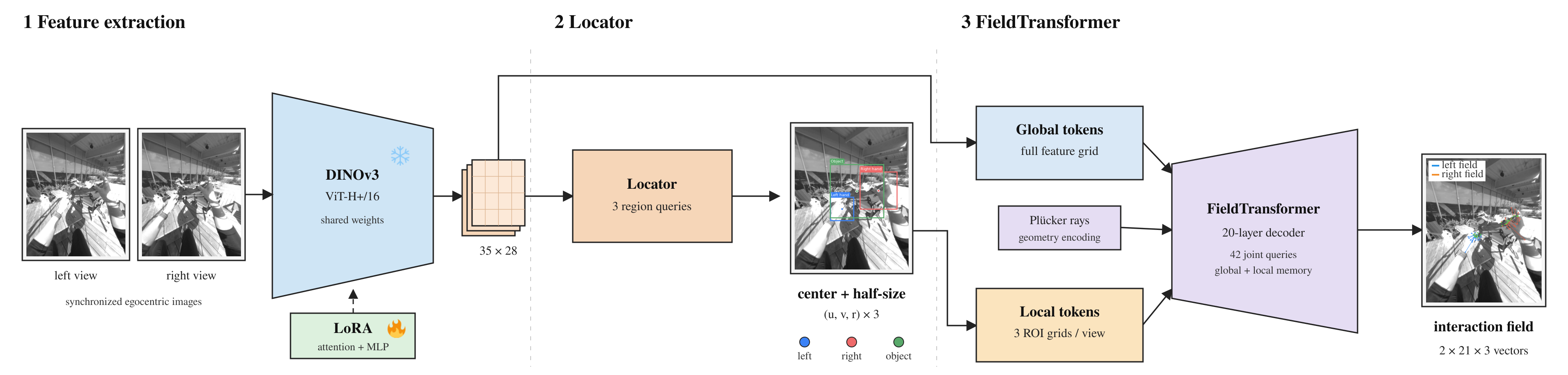}
    \caption{Overview of \method. A shared LoRA-adapted DINOv3 encoder processes synchronized headset views. The locator predicts square hand/object regions, and the FieldTransformer decodes 42 joint vectors from global tokens, RoI tokens, and calibrated Pl\"ucker-ray embeddings.}
    \label{fig:overview}
\end{figure*}

\method{} combines three ideas (\cref{fig:overview}).
First, densely sampled square hand/object RoIs emphasize local detail without discarding dense context.
Second, reference-frame Pl\"ucker embeddings expose calibrated stereo geometry.
Third, DINOv3 ViT-H+/16~\cite{simeoni2026dinov3} is efficiently adapted with LoRA~\cite{hu2022lora} and \mbox{LayerNorm} affine parameters.
A query decoder jointly predicts both interaction fields.

\section{Method}

\subsection{Interaction Field}

For hand $h\in\{L,R\}$ and joint $j$, let $\mathbf p_{h,j}$ denote the 3D joint position and $\mathbf p^*_{h,j}$ its closest point on the manipulated object surface.
The interaction-field target is
\begin{equation}
    \mathbf v_{h,j}=\mathbf p^*_{h,j}-\mathbf p_{h,j}\in\mathbb R^3.
\end{equation}
Given stereo images and calibration, the network predicts $\hat{\mathbf V}\in\mathbb R^{2\times21\times3}$ in the reference-camera frame and minimizes masked mean ADE over valid hands. Before evaluation or submission, each prediction is rotated by the per-frame calibration $R_{\mathrm{world}\leftarrow\mathrm{ref}}$ into the required world frame; all vectors remain in millimeters.

\subsection{Global--Local Representation}

Both views share a DINOv3 ViT-H+/16 encoder~\cite{simeoni2026dinov3}.
Original backbone weights are frozen except for \mbox{LayerNorm} affine parameters, while LoRA adapters are trained.
For a pretrained projection $W_0$, LoRA applies
\begin{equation}
    y=W_0x+\frac{\alpha}{r}BAx.
\end{equation}
Rank-128 LoRA with $\alpha=256$ adapts attention QKV/output and all three SwiGLU projections in every block.

For each view $v$, the field branch fuses blocks $\{10,20,26,31\}$ with learned softmax weights into feature map $F^v$; the locator consumes block 31 directly.
Its eight-layer, three-query decoder predicts center $\mathbf c_t^v$, square half-size $s_t^v$, and presence for each hand and the object.
The object-region query is additionally conditioned on a learned embedding of the provided object category.
Center and log-size use Smooth L1 loss, presence uses binary cross-entropy, and coverage measures the target extent left outside the predicted square, normalized by 64 pixels. Their weights are $1,1,1,$ and $0.1$, respectively; missing hands are masked from the regression terms.

\begin{figure}[t]
    \centering
    \includegraphics[width=0.98\columnwidth]{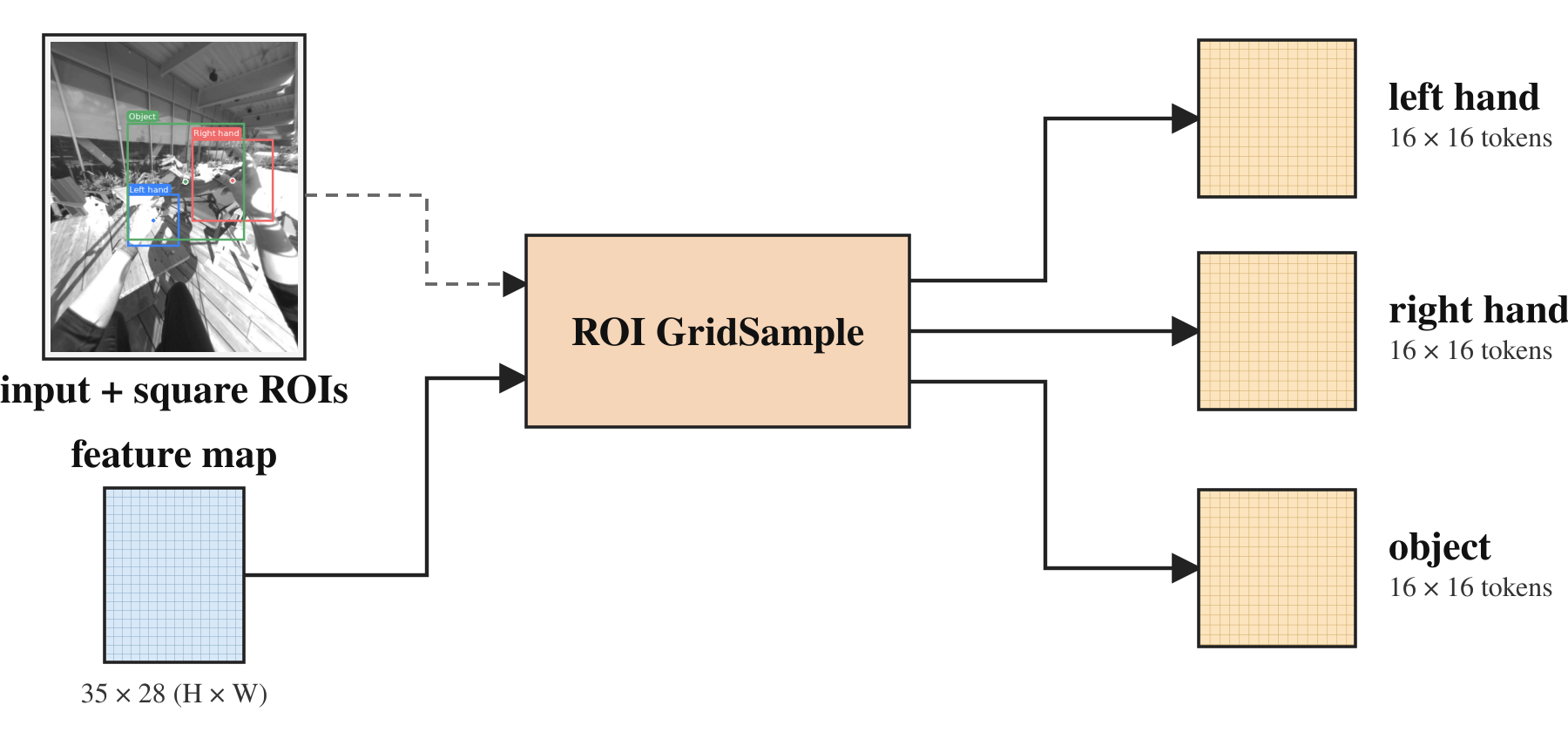}
    \caption{Local-token construction. The portrait $35\times28$ ($H\times W$) feature map is sampled at the locator grids to form three $16\times16$ RoI token maps.}
    \label{fig:local_tokens}
\end{figure}

For each predicted region, a regular $16\times16$ grid $G$ bilinearly samples spatially aligned local RoI tokens:
\begin{equation}
R_t^v=\operatorname{GridSample}\left(F^v,\mathbf c_t^v+s_t^vG\right),
\quad t\in\{L,R,O\}.
\end{equation}
Role embeddings distinguish the three regions, while the full $35\times28$ feature map is retained as dense global context (\cref{fig:local_tokens}).

\subsection{Calibration-Aware Stereo Decoder}

We back-project token pixel $\mathbf u$ in camera $k$ with $(K^k)^{-1}[u,v,1]^\top$ and transform ray direction $\mathbf d$ and origin $\mathbf o$ to the reference frame. We normalize $\mathbf d$ to unit length and express $\mathbf o$ in meters before forming the moment.
The ray is encoded using origin-aware Pl\"ucker coordinates~\cite{sitzmann2021lfn}
\begin{equation}
    \boldsymbol\pi(\mathbf u,k)=[\mathbf d,\mathbf o\times\mathbf d]\in\mathbb R^6.
\end{equation}
A two-layer MLP maps rays to decoder width; ray and view embeddings are added to all tokens.
Global rays use patch centers and local rays use actual GridSample locations.
Reference-view swapping discourages camera-role shortcuts, and memory concatenates global and three local RoI token sets from both views.
We adopt Joint Transformer's joint-indexed query formulation~\cite{knaebel2023jointtransformer}: 42 learned queries---one per hand--joint pair---attend to our global--local stereo memory through a 20-layer Transformer with width 512 and eight heads.
A shared linear head outputs each query's 3D vector.

\subsection{Training}

All stages use AdamW with cosine decay. We train the monocular locator and LoRA for 20 epochs with total batch 32 and learning rate $10^{-4}$, treating both cameras independently and using photometric/noise augmentation, HFlip, partial crop, and $\pm20^\circ$ rotation. Weight decay is $5\times10^{-4}$ for the locator and zero for LoRA and \mbox{LayerNorm} parameters.
The 40-epoch stereo stage uses total batch 96, learning rates of $5\times10^{-5}$ for the FieldTransformer and $10^{-4}$ for LoRA, and weight decay $5\times10^{-4}$ only on the FieldTransformer, with shared $\pm10^\circ$ rotation, photometric jitter, HFlip, and reference swapping.
The locator decoder and heads remain frozen; its auxiliary loss still updates the shared backbone adapters, whereas field gradients stop at the predicted RoI coordinates.
We then enable \mbox{LayerNorm} affine parameters and jointly fine-tune them, LoRA, and the FieldTransformer for 20 epochs at $10^{-4}$.
After fixing all choices, we return the two development subjects to the official training set, keep the locator frozen, and fine-tune the FieldTransformer, LoRA adapters, and \mbox{LayerNorm} affine parameters for two epochs at $2\times10^{-5}$.
For horizontal-flip test-time augmentation (HFlip TTA), both views and their intrinsics are mirrored together. We restore the prediction by swapping left/right slots and negating the camera-frame $x$ component, rotate both branches to the world frame, and average them.

\section{Experiments}

\paragraph{Setup.}
The SHOW3D training set contains 468 recordings from 10 subjects and 21 objects, sampled at 10 FPS.
The hidden test set contains 20,042 frames sampled at 5 FPS from 139 recordings, covering three unseen subjects and 16 objects.
For development, we hold out subjects \texttt{XYZ109} and \texttt{LYA722} as a clean validation split.
Inputs are resized to $560\times448$ ($H\times W$).
The server reports left/right ADE, mean ADE, recall, and an official ranking score. The official score is a hidden server-side metric that jointly accounts for prediction accuracy and coverage; its exact formula is not public. All error metrics and the official score are lower-is-better.

\paragraph{Input resolution.}
During development, we use a matched monocular DINOv3 ViT-L/16 setup to tune the input resolution (\cref{tab:resolution}).
These runs are used only for hyperparameter selection and differ from the final model evaluated on Codabench, so we report validation results only.

\begin{table}[t]
\centering
\caption{Development-set resolution tuning under a matched monocular DINOv3 ViT-L/16 setting. Lower is better.}
\label{tab:resolution}
\footnotesize
\setlength{\tabcolsep}{8.0pt}
\begin{tabular}{@{}lc@{}}
\toprule
Resolution ($H\times W$) & Val mean ADE (mm) \\
\midrule
$400\times320$ & 44.22 \\
$480\times384$ & 41.38 \\
$560\times448$ & \textbf{40.23} \\
$640\times512$ & 40.51 \\
\bottomrule
\end{tabular}
\end{table}

\paragraph{Feature-space versus image-space crops.}
Under a matched configuration, RGB-crop re-encoding and feature-space RoI sampling achieve comparable validation mean ADE (37.93 versus 37.77~mm).
This setting differs from other experiments and supports only this paired comparison.
We choose feature-space sampling for its single backbone pass and lower cost.

\paragraph{Backbone adaptation.}
We compare staged adapters on the clean split (\cref{tab:adapter}).
Larger attention/MLP ranks correlate with lower errors; LayerNorm affine adaptation gives the best run at 10.04 pixels and 28.31/28.04~mm raw/HFlip-TTA mean ADE.
As the rows come from successive stages, they indicate capacity effects rather than a strict one-variable ablation.

\begin{table}[t]
\centering
\caption{Backbone adaptation on the development split. Raw uses the original stereo pair; HFlip TTA averages the raw and coordinate-restored flipped predictions. Both ADE columns report validation mean ADE in millimeters; lower is better.}
\label{tab:adapter}
\footnotesize
\setlength{\tabcolsep}{2.5pt}
\begin{tabular}{@{}lccc@{}}
\toprule
Backbone adaptation & Center err. (px) & Raw ADE & HFlip-TTA ADE \\
\midrule
Attn 64 + MLP 16 & 10.46 & 29.11 & 28.84 \\
Attn 128 + MLP 128 & 10.12 & 28.82 & 28.41 \\
$+$ LayerNorm affine & \textbf{10.04} & \textbf{28.31} & \textbf{28.04} \\
\bottomrule
\end{tabular}
\end{table}

\paragraph{Challenge results.}
\Cref{tab:results} traces submissions rather than controlled ablations.
Stereo fusion improves performance, and reference-frame Pl\"ucker encoding further lowers mean ADE from 36.17 to 35.57~mm; broader ViT-H+ adaptation closes most of the remaining gap.

\begin{table}[t]
\centering
\caption{Submission progression on the hidden SHOW3D test set. Lower is better.}
\label{tab:results}
\footnotesize
\setlength{\tabcolsep}{3.2pt}
\begin{tabular}{@{}lcc@{}}
\toprule
System & Official & Mean ADE (mm) \\
\midrule
InterField, ResNet-50 & 58.98 & 59.38 \\
DINOv3 ViT-L/16, one view, $400\times320$ & 43.12 & 43.50 \\
$+$ both-view training/HFlip, $560\times448$ & 38.74 & 39.10 \\
$+$ local RoI tokens & 36.01 & 36.36 \\
$+$ stereo fusion & 35.83 & 36.17 \\
$+$ reference-frame Pl\"ucker encoding & 35.21 & 35.57 \\
ViT-H+, LoRA 64/16 & 29.51 & 29.86 \\
ViT-H+, LoRA 128/128 + LN & 28.55 & 28.91 \\
$+$ full-training-set fine-tuning & 27.61 & 27.96 \\
$+$ direct-FT last-24-block ensemble & \textbf{27.47} & \textbf{27.82} \\
\bottomrule
\end{tabular}
\end{table}

The primary model reaches 27.61 official score and 27.96~mm mean ADE; equal-weight averaging with a last-24-block fine-tuned variant reaches 27.47 and 27.82~mm, respectively, ranking first.

\section{Conclusion}

Combining global--local features, stereo geometry, and efficient backbone adaptation, GOLF's final ensemble scores 27.47 and places first in the HANDS@ECCV 2026 SHOW3D Challenge.

{\small
\setlength{\bibsep}{0pt}
\bibliographystyle{ieeenat_fullname}
\bibliography{references}

@inproceedings{banerjee2025hot3d,
  title     = {{HOT3D}: Hand and Object Tracking in {3D} from Egocentric Multi-View Videos},
  author    = {Banerjee, Prithviraj and Shkodrani, Sindi and Moulon, Pierre and others},
  booktitle = {CVPR},
  pages     = {7061--7071},
  year      = {2025}
}

@inproceedings{fan2023arctic,
  title     = {{ARCTIC}: A Dataset for Dexterous Bimanual Hand-Object Manipulation},
  author    = {Fan, Zicong and Taheri, Omid and Tzionas, Dimitrios and others},
  booktitle = {CVPR},
  pages     = {12943--12954},
  year      = {2023}
}

@inproceedings{hu2022lora,
  title     = {{LoRA}: Low-Rank Adaptation of Large Language Models},
  author    = {Hu, Edward J. and Shen, Yelong and Wallis, Phillip and others},
  booktitle = {ICLR},
  year      = {2022}
}

@misc{knaebel2023jointtransformer,
  title     = {{Joint Transformer}},
  author    = {Abou Zeid, Karim},
  year      = {2023},
  howpublished = {HANDS@ICCVW},
  note      = {\href{https://github.com/karimknaebel/JointTransformer}{GitHub repository}}
}

@inproceedings{rim2026show3d,
  title     = {{SHOW3D}: Capturing Scenes of {3D} Hands and Objects in the Wild},
  author    = {Rim, Patrick and Harris, Kevin and Copple, Braden and others},
  booktitle = {CVPR},
  pages     = {7111--7120},
  year      = {2026}
}

@article{simeoni2026dinov3,
  title   = {{DINOv3}},
  author  = {Sim{\'e}oni, Oriane and Vo, Huy V. and Seitzer, Maximilian and others},
  journal = {Transactions on Machine Learning Research},
  year    = {2026}
}

@inproceedings{sitzmann2021lfn,
  title     = {Light Field Networks: Neural Scene Representations with Single-Evaluation Rendering},
  author    = {Sitzmann, Vincent and Rezchikov, Semon and Freeman, William T. and others},
  booktitle = {NeurIPS},
  volume    = {34},
  pages     = {19313--19325},
  year      = {2021}
}
}

\end{document}